\documentclass[3p,times,procedia]{elsarticle}
\usepackage{ecrc}
\usepackage[bookmarks=false]{hyperref}
    \hypersetup{colorlinks,
      linkcolor=blue,
      citecolor=blue,
      urlcolor=blue}

\volume{00}

\firstpage{1}

\journalname{Procedia Computer Science}

\runauth{T. Chiaburu, F. Bießmann}

\jid{procs}

\usepackage{amssymb}
\usepackage{amsthm}
\usepackage{subfig}
\usepackage{listings}
\usepackage{array}

\begin{document}
\begin{frontmatter}



\dochead{International Conference on Industry Sciences and Computer Science Innovation}

\title{Interpretable Time Series Models for Wastewater Modeling in Combined Sewer Overflows}


\author[1]{Teodor Chiaburu\corref{cor1}} 
\author[1,2]{Felix Bießmann}

\address[1]{Berliner Hochschule für Technik, Luxemburger Str. 10, 13353 Berlin, Germany}
\address[2]{Einstein Center Digital Future, Wilhelmstraße 67, 10117 Berlin, Germany}

\begin{abstract}
Climate change poses increasingly complex challenges to our society. Extreme weather events such as floods, wild fires or droughts are becoming more frequent, spontaneous and difficult to foresee or counteract. In this work we specifically address the problem of sewage water polluting surface water bodies after spilling over from rain tanks as a consequence of heavy rain events. We investigate to what extent state-of-the-art interpretable time series models can help predict such critical water level points, so that the excess can promptly be redistributed across the sewage network. Our results indicate that modern time series models can contribute to better waste water management and prevention of environmental pollution from sewer systems. All the code and experiments can be found in our repository: \href{https://github.com/TeodorChiaburu/RIWWER_TimeSeries}{https://github.com/TeodorChiaburu/RIWWER\_TimeSeries}.
\end{abstract}

\cortext[cor1]{Corresponding author.}
\end{frontmatter}

\email{chiaburu.teodor@bht-berlin.de}



\section{Introduction}

\begin{figure}
  \centering
  \includegraphics[width=\linewidth]{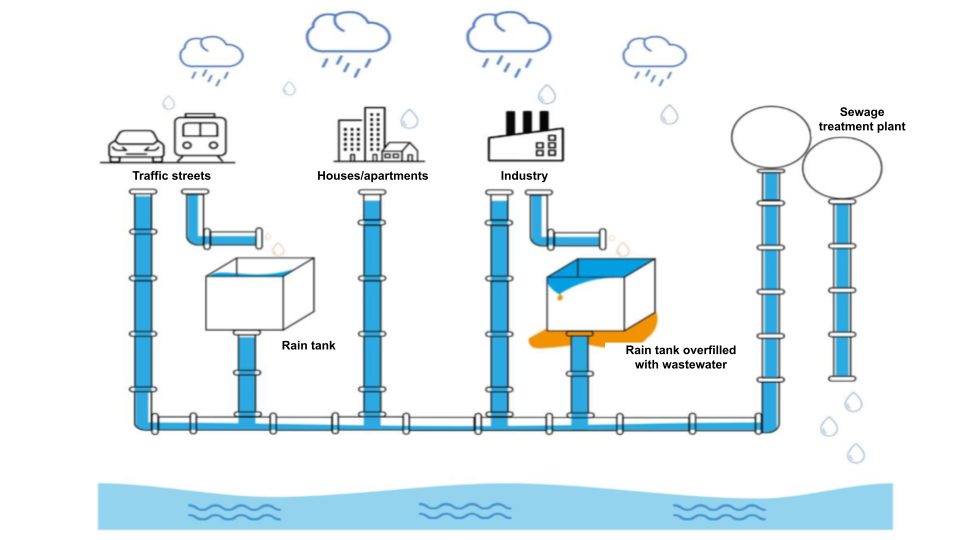}
  \caption{The application scenario of a hydrological network with combined rain and wastewater sewer overflows: Rain water is collected from different parts of the city in multiple rain tanks. On a heavy rain event, these tanks may get filled above capacity, spilling the wastewater into the environment.}
  \label{fig:hydro_network}
\end{figure}

Extreme rainfall events have become more frequent in the last decades \cite{roxyThreefoldRiseWidespread2017}. Heavy rainfalls can have drastic consequences on wastewater infrastructure and cause sudden \textit{hydraulic stress} on sewer systems that combine wastewater and rain water. In such extreme rainfall events Combined Sewer Overflow (CSO) systems release wastewater into the environment (see Figure \ref{fig:hydro_network}). This can become dangerous if the volumes exceed the natural capacity of the ecosystem to handle it. Several negative impacts include: contamination of water bodies with pollutants - posing health risks for human consumption and aquatic organisms, depletion of oxygen levels in the water, soil contamination and more. In 2021, the German city of Ahr was faced with such a catastrophe that left traces till today (German article: \href{https://reportage.wdr.de/chronik-ahrtal-hochwasser-katastrophe}{https://reportage.wdr.de/chronik-ahrtal-hochwasser-katastrophe}).

In order to minimize the wastewater spillover, sewage networks need to be adjusted to the current environmental challenges. One important aspect is the ability to forecast the behaviour or state of complex sewage systems. Recently, Machine Learning (ML) methods have been successfully used for modelling sewer systems \cite{loweReviewMachineLearning2022,zhuReviewApplicationMachine2022} and comprehensive benchmark data sets were published to foster research in the domain of wastewater management \cite{bellinge}. Building on this work, we investigate here the potential of recently developed state-of-the-art time series (TS) models in the context of wastewater systems. 

\section{Overview of the Data}

The dataset for the following experiments was gathered by \textit{Wirtschaftsbetriebe Duisburg} (\href{https://www.wb-duisburg.de/}{website}\footnote{https://www.wb-duisburg.de/} in German). It comprises 20 sensors distributed along the sewage system in five neighborhoods in Duisburg, one rain data column and the wastewater levels for five rain collecting tanks\footnote{The dataset is currently nor freely available online.}. Since the sensors are programmed in an event based fashion, data was recorded only when there was a change in the sensor value. To ensure comparability to standard datasets \cite{bellinge}, we resampled our data hourly and averaged out sensor values whenever there were more recordings within one hour. This led to a total of 8760 samples, starting from the 1st of January 2021 at midnight up until the 31st of December 2021 at 23:00.

In order to model and forecast overflow events it is necessary to accurately and robustly predict the volume of wastewater in rain tanks. Therefore, we define the prediction problem in this study as a time series prediction task with the volume in rain tanks as the target variable and all other sensors as exogenous input variables.

In order to avoid leakage of training data into the test data set we follow the usual time series cross-validation split procedure: the first 80\% samples (chronologically) of the data were assigned as training set, the next 20\% as validation set. 
The window length of the data ingested by the encoder of our TS models and the forecasting horizon for the predictions are two of the most important hyperparameters in this setting; our hyper parameter optimization (HPO) procedures set the encoder length to 48 hours and the prediction span (or \textit{horizon}) to 10 hours. Any  prediction beyond 10 hours turned out to be highly inaccurate.

\section{Methods}

Given the temporal structure of the sensor data, a common approach is to use TS models to predict the wastewater levels in the rain tanks, which prove to be superior to other modelling strategies such as CNNs or GANs \cite{floodGAN}. We ran our experiments with two state-of-the-art (SoTA) TS models from the \textit{PyTorch-Forecasting} \href{https://pytorch-forecasting.readthedocs.io/en/stable/}{library}\footnote{https://pytorch-forecasting.readthedocs.io/en/stable/}, namely N-HiTS\footnote{We note that the library also implements the model's predecessor - N-Beats \cite{nbeats}. However, in contrast to its more modern version, N-Beats cannot handle covariates, which is a design necessity for our scenarios where a wide array of sensors would deliver the complex input for the model.} \cite{nhits} and TFT \cite{tft}. One of their advantages is that their predictions are more interpretable than those of classical TS models. Interpretability is a valuable feature in real life scenarios where water engineers need to understand AI-generated predictions for critical levels of wastewater, since emptying the rain collecting tanks in due time is an expensive process.

N-HiTS captures patterns in the time sequence by projecting the signal onto a basis of simple functions, such as sine waves or step functions. The basis functions that make up the prediction (at multiple resolutions) are then visualized as explanations for the model's decision (see Fig. \ref{fig:nhits_interp}). The TFT applies recurrent layers enhanced with interpretable self-attention modules. Since the introduction of attention mechanisms for translation \cite{att_text}, the method has been adapted to other fields as well, such as image classification \cite{att_img} or tabular learning \cite{att_tab} and in recent years for temporal data as well.

The next section describes the results of our experiments with the two SoTA models on the German dataset. We also draw a parallel to how well the winning model performs on the standard sewage dataset from the literature, namely the Danish Bellinge dataset \cite{bellinge}. More details about the training routines and hypertuning can be read in the Supplement.

\section{Results}

As described above, we have trained two modern interpretable TS models on two data sets, one from Duisburg, Germany, and a standard benchmark data set from Bellinge \cite{bellinge}. We tested for both of them two training routines: one with the default loss Mean Average Scaled Error (MASE) and one extending the network from a point-forecaster to a distribution predictor via the Quantile Loss (Q-Loss)\footnote{We used (0.02, 0.1, 0.25, 0.5, 0.75, 0.9, 0.98)-quantiles.}. For comparison purposes, the baseline TS model from the library was also applied (its forecasts are based on simple linear extrapolations from the last known backcast values) - see Table \ref{table:val_metrics}. It turns out that N-HiTS trained on MASE performs best on the given dataset. It is also the fastest predictor, as far as inference latency is concerned, but not the smallest though; however, PyTorch offers several \href{https://pytorch.org/docs/stable/quantization.html}{quantization techniques}\footnote{https://pytorch.org/docs/stable/quantization.html} for compressing models at minimal performance loss. Apart from accurate predictions, size and inference speed also play an important role in the model selection for our use cases, since the goal is to run these predictors on edge devices installed across the sewage network.

\subsection{Duisburg Dataset Results}

\begin{table}[h!]
    \centering
    \begin{tabular}{|m{6em} || m{4em} | m{4em} | m{4em} | m{4em} | m{4em}|} 
     \hline
     \textbf{\textit{Metric}} & \textbf{Baseline} & \textbf{N-HiTS (MASE)} & \textbf{N-HiTS (Q-Loss)} & \textbf{TFT (MASE)} & \textbf{TFT \quad (Q-Loss)} \\
     \hline\hline
     MAE & 1.7076 & \textbf{1.4225} & 1.7085 & 1.6755 & 2.6277 \\ 
      \hline
     RMSE & 10.2045 & \textbf{3.2073} & 3.4515 & 3.4223 & 4.2043 \\
     \hline
     Size[Mb] & - & 17.9 & 18.9 & \textbf{0.8} & 1.1 \\
     \hline
     Latency [ms/sample] & - & \textbf{1.0940} & 1.1619 & 1.2993 & 1.4842 \\
     \hline
    \end{tabular}
    \caption{Performance scores for the two SoTA models and a simple baseline trained on the Duisburg dataset. The Mean Average Error (MAE) and the Root Mean Squared Error (RMSE) are calculated on the validation set.}
    \label{table:val_metrics}
\end{table}

\begin{figure}[h]
    \centering
    \subfloat[N-HiTS MASE Loss]{\includegraphics[width=0.45\textwidth]{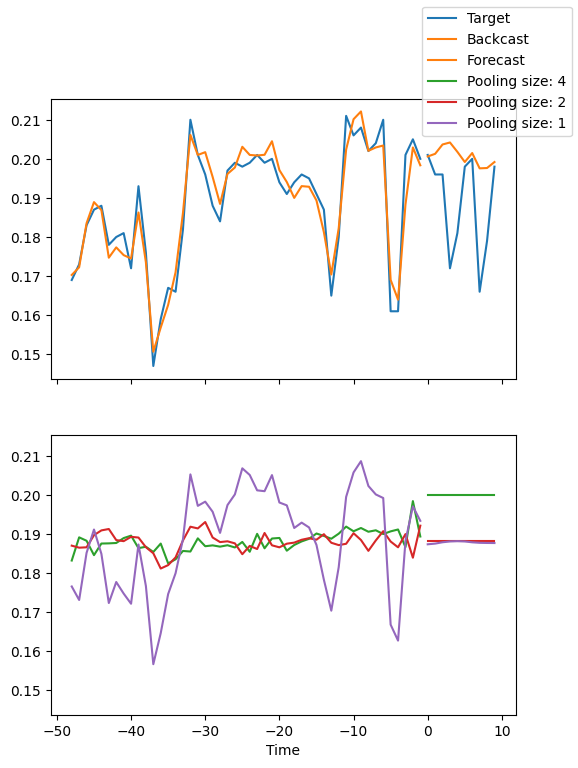}}\hfil
    \subfloat[N-HiTS Quantile Loss]{\includegraphics[width=0.45\textwidth]{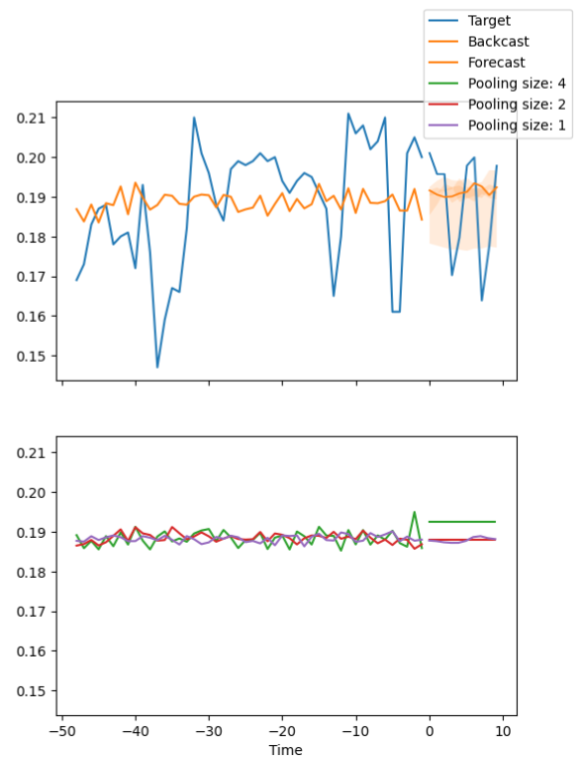}}\hfil 
    
    \caption{Predictions of the normalized wastewater volumes (top row) and explanations (bottom row) for the Duisburg dataset with N-HiTS. The purple, green and red lines are additive components of the model. Note that the purple component (for pooling size 1) in the left column predominantly captures the signal in the data.}
    \label{fig:nhits_interp}
\end{figure}

\begin{figure}[h]
    \centering
    \subfloat[TFT MASE Loss]{\includegraphics[width=0.45\textwidth]{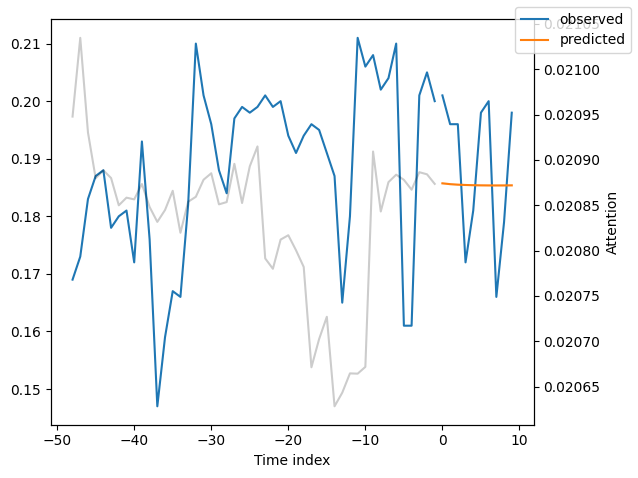}}\hfil
    \subfloat[TFT Quantile Loss]{\includegraphics[width=0.45\textwidth]{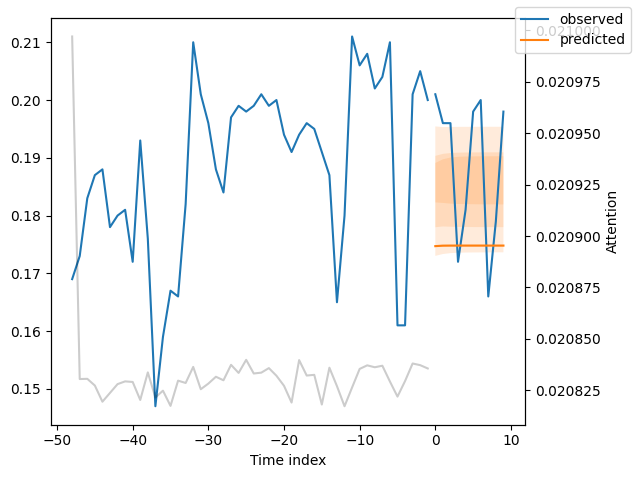}}\hfil 
    
    \caption{Predictions of the normalized wastewater volumes for the Duisburg dataset with TFT. The gray curve highlights the attention scores for the corresponding point in time.}
    \label{fig:tft_pred}
\end{figure}

\begin{figure}[h]
    \centering
    \subfloat[TFT MASE Loss]{\includegraphics[width=0.45\textwidth]{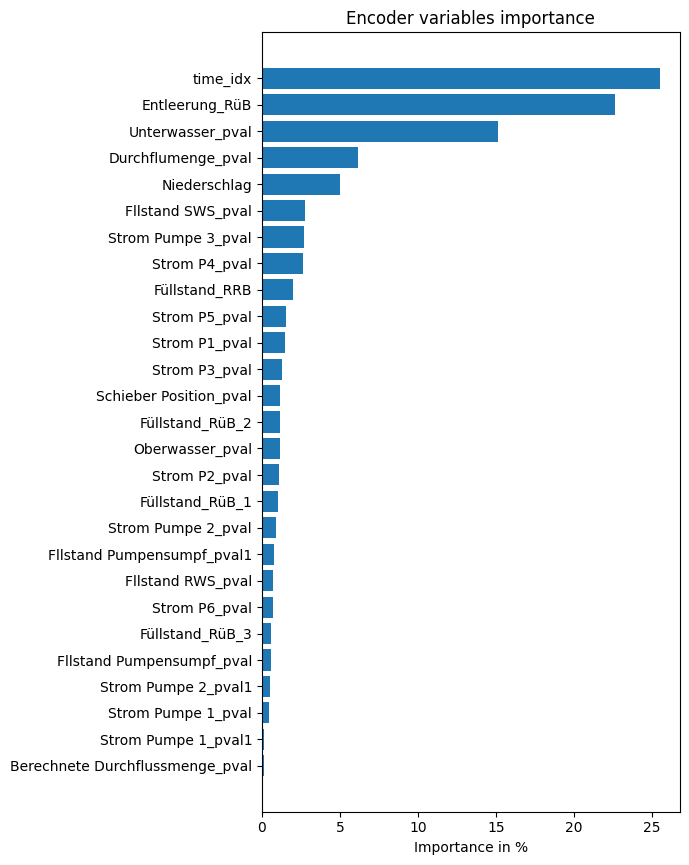}}\hfil
    \subfloat[TFT Quantile Loss]{\includegraphics[width=0.45\textwidth]{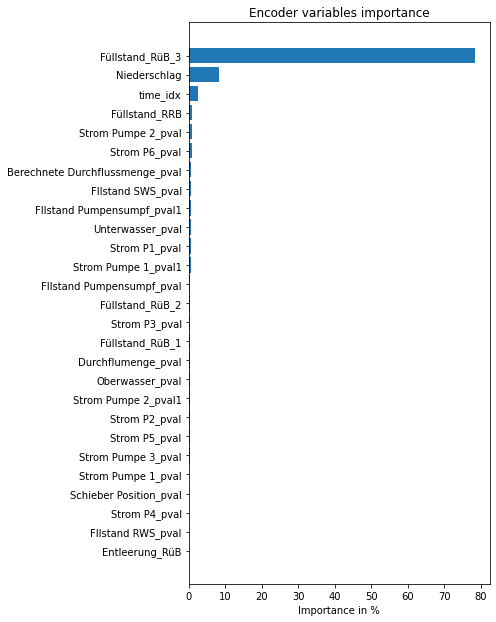}}\hfil 
    
    \caption{Feature importances for the two versions of TFT on the Duisburg dataset.}
    \label{fig:tft_interp}
\end{figure}

Figures \ref{fig:nhits_interp} and \ref{fig:tft_pred} show the same sample from the German dataset being analyzed by the four model configurations (more examples in the notebooks in our repository). N-HiTS with MASE appears to have learnt the temporal structure in the backcast phase quite well. The fact is confirmed by the basis functions the backcast predictions is composed of; with increasing resolutions (pooling sizes), the projections on the basis functions seem to successfully capture different levels of detail in the signal. For the forecast, there is room for improvement though. The Q-Loss version of N-HiTS does make a more realistic prediction for the future; yet, the backcast patterns are not as convincing as the ones in the lower left figure. The TFT delivers in both cases rather coarse predictions; we learn from the attention curves (highlighted in gray) that the model seems to be focusing predominantly on the first several hours of the signal, discarding part or most of the information closer to the forecast block. Figure \ref{fig:tft_interp} reveals the feature importances computed from the averaged attention scores in the validation set for the two TFTs. The Q-Loss TFT is obviously neglecting most of the features, concentrating only on one rain tank, the weather data and the time index. Its counterpart does make use of much more information, but even here we notice a skewed distribution towards only a handful of features.

\subsection{Bellinge Dataset Results}

\begin{figure}[h]
    \centering
    \subfloat[]{\includegraphics[width=0.45\textwidth]{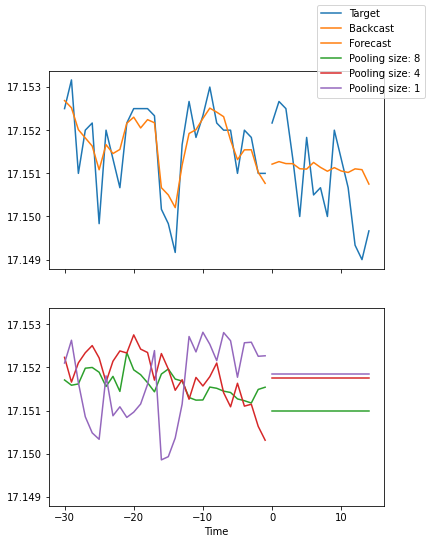}}\hfil
    \subfloat[]{\includegraphics[width=0.45\textwidth]{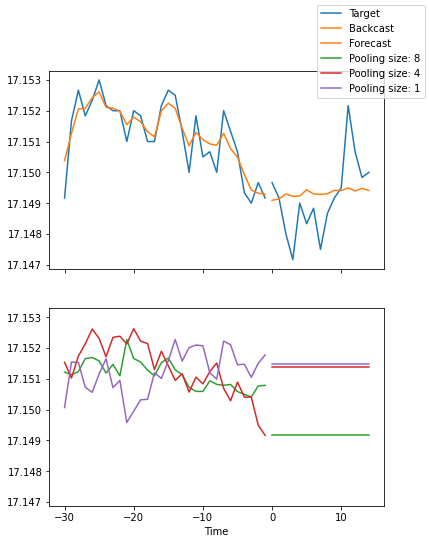}}\hfil 
    
    \caption{Predictions of the wastewater level [cm] (top row) and explanations (bottom row) with N-HiTS (MASE) for two samples from the Bellinge dataset, subset for sensor G71F04R.}
    \label{fig:bel}
\end{figure}

The most well known dataset for urban drainage systems is the Bellinge dataset, gathered by the Danish authorities in the city of Odense. It is very well curated and documented. It contains a wide range of sensors and weather radars, recording data for the past 10 years. We tested our N-HiTS configuration with MASE on one of the Bellinge sensors, namely on \textit{G71F04R}\footnote{We note that the sensor is monitoring one specific rain tank and that data is recorded every minute, not on an hourly basis like in the Duisburg setting. The data sheet for this sensor has over 54 k samples.}. We report an MAE of 0.0035 and RMSE of 0.0047 on the validation set\footnote{Again, the training-validation ratio was 80-20. The backcast block encompassed 30 minutes, the forecast 15 minutes.}, which is an obvious performance improvement compared to the Duisburg scores. Two example predictions can be seen in Fig. \ref{fig:bel} from the Supplement.

\subsection{Simulating Sensor Failure}

One expected scenario is that one or multiple sensors along the sewer system fail, which means they stop recording data at all or record faulty values. Apart from high performance, speed, low energy consumption and interpretability, our model is also expected to be robust against corrupted features. In order to simulate this case, we have iteratively 'shut down' sensors from the Duisburg dataset and looked at how the N-HiTS performance on the validation set changes. The sensors in Duisburg are clustered into five quarters across the city; it turned out that it is enough to corrupt one single sensor cluster and the prediction errors (MAE and RMSE) would increase by a factor of 2 to 4. For instance, when corrupting the five sensors in the Herzogstr. cluster in Duisburg, the MAE increased to 3.1364 and the RMSE to 11.1946. The corruption was done by picking a random percentile of the sensor's value distribution between 0 - 10\% and setting the following 90\% of the unique values to 0 (example in the notebook \textit{NHits\_Vierlinden\_Dropout.ipynb} from the repository).

\section{Conclusion}

To summarise, modern interpretable TS models could offer a viable solution for tackling wastewater level prediction as a consequence of heavy rain events. It remains, however, challenging to deal with the requirements - on both sides of the software and hardware needed - of such a complex system. Our preliminary results indicate that the robustness of modern TS models is not satisfactory when there are sensor outages. This highlights the need for more research on methods dedicated to this application scenario. Furthermore, our methods are only predictive models that do not account for actors in the system. For practical applications of ML models in automated sewer systems one will need to embed such TS models as components of a Reinforcement Learning model, similar to the approaches in \cite{swmm, rl_sewage, bellinge, accel_phys}.

\section*{Acknowledgments}
  The authors acknowledge funding by the Federal Ministry of Economic Affairs and Climate Action through the research project RIWWER – "Reduction of the Impact of untreated Waste Water on the Environment in case of torrential Rain" (01MD22007H).

\section*{Supplementary Material}

\subsection*{Training Procedure}

The hyperparameters of the TFT and N-HiTS have been optimized with the \textit{optuna} \href{https://optuna.org/}{library}\footnote{https://optuna.org/}. The results are summarized in Table \ref{table:hp}. For training all models, \textit{Stochastic Weight Averaging} \cite{swa} was used as learning rate schedule and an Early Stopper.
\vspace{2cm}

\begin{table}[h!]
    \centering
    \begin{tabular}{|m{7.1em} || m{4em} | m{4em} | m{4em} | m{4em} | m{4em}| m{4em}| m{4em}| m{4em}|} 
     \hline
     \textbf{Hyperparams. / Models} & \textbf{LR} & \textbf{Dropout} & \textbf{Weight Decay} & \textbf{Grad. Clipping} & \textbf{Hidden Size} & \textbf{Attention Head Size} & \textbf{Backcast Loss \quad Ratio} \\
     \hline\hline
     TFT (MASE) & 0.0151 & 0.2 & 1e-3 & 0.2036 & 4 & 3 & -\\ 
      \hline
     TFT (Q-Loss) & 0.0138 & 0.1292 & 1e-3 & 0.2961 & 8 & 4 & - \\
     \hline
     N-HiTS (MASE) & 0.0126 & 0.1 & 1e-2 & 0.01 & 512 & - & 1.0 \\
     \hline
     N-HiTS (Q-Loss) & 3.98e-5 & 0.1 & 1e-2 & 0.01 & 512 & - & 0.0 \\
     \hline
    \end{tabular}
    \caption{Hyperparameters for all the four models.}
    \label{table:hp}
\end{table}

\subsection*{Inference on Bellinge Dataset}

\pagebreak
\bibliography{riwwer_bib}
\bibliographystyle{elsarticle-harv}

\end{document}